\documentclass[akbc,twoside,11pt,lettersize]{article}
\usepackage{akbc}
\akbcheading
\ShortHeadings{CascadER: Cross-Modal Cascading for Knowledge Graph Link Prediction}{Safavi, Downey, Hope}
\finalcopy % Uncomment for camera-ready version, but NOT for submission.

\usepackage[utf8]{inputenc} % allow utf-8 input
\usepackage[T1]{fontenc}    % use 8-bit T1 fonts
\usepackage{hyperref}       % hyperlinks
\usepackage{url}            % simple URL typesetting
\usepackage{booktabs}       % professional-quality tables
\usepackage{amsfonts}       % blackboard math symbols
\usepackage{nicefrac}       % compact symbols for 1/2, etc.
\usepackage{microtype}      % microtypography
\usepackage{xcolor}         % colors
\usepackage{graphicx}
\usepackage{amsmath}
\usepackage{amssymb}
\usepackage{bbm}
\usepackage{multicol}
\usepackage{multirow}
\usepackage{subcaption} 
\usepackage{amsmath}
\usepackage{floatrow}
\newfloatcommand{capbtabbox}{table}[][\FBwidth]

\newcommand{\method}{\textsc{CascadER}}
\newcommand{\cascader}{\textsc{CascadER}}
\newcommand{\kg}{KG}
\newcommand{\kge}{KGE}
\newcommand{\kb}{\kg}
\newcommand{\lm}{LM}

\newcommand{\repodb}{\textsc{RepoDB}}
\newcommand{\codex}{\textsc{CoDEx}}
\newcommand{\codexS}{\textsc{CoDEx-S}}
\newcommand{\codexM}{\textsc{CoDEx-M}}
\newcommand{\fb}{\textsc{FB15K-237}}
\newcommand{\wn}{\textsc{WN18RR}}

\newcommand{\CLS}{\texttt{[CLS]}}
\newcommand{\SEP}{\texttt{[SEP]}}

\newcommand{\G}{\mathcal{G}}
\newcommand{\V}{\mathcal{E}}

\newcommand{\relations}{\mathcal{R}}

\newcommand{\firstRanker}{f^{(t)}}
\newcommand{\secondRanker}{f^{(t + 1)}}
\newcommand{\numTest}{N_{\textrm{test}}}

\newcommand{\scoreMatrix}{\mathbf{S}}

\newcommand{\realNumbers}{\mathbb{R}}

\DeclareMathOperator*{\argmax}{arg\,max}

\newcommand{\loss}{\mathcal{L}}

\begin{document}

\title{CascadER: Cross-Modal Cascading for \\ Knowledge Graph Link Prediction}

\author{%
\centering
  Tara Safavi$^{1}$\thanks{Work performed during an internship at the Allen Institute for Artificial Intelligence.},~~ Doug Downey$^{2,3}$,~~ Tom Hope$^{2,4}$ \\
  \texttt{tsafavi@umich.edu}, \texttt{dougd@allenai.org}, \texttt{tomh@allenai.org} \\
  $^{1}$University of Michigan, $^{2}$Allen Institute for Artificial Intelligence (AI2),\\
  $^{3}$Northwestern University, $^{4}$The Hebrew University of Jerusalem \\ 
  % \AND
  % Coauthor \\
  % Affiliation \\
  % Address \\
  % \texttt{email} \\
  % \And
  % Coauthor \\
  % Affiliation \\
  % Address \\
  % \texttt{email} \\
  % \And
  % Coauthor \\
  % Affiliation \\
  % Address \\
  % \texttt{email} \\
}

\maketitle

\begin{abstract}
Knowledge graph (\kg{}) link prediction is a fundamental task in artificial intelligence, with applications in natural language processing, information retrieval, and biomedicine. 
Recently, promising results have been achieved by leveraging cross-modal information in \kg{}s, using ensembles that combine knowledge graph embeddings (\kge{}s) and contextual language models (\lm{}s). 
However, existing ensembles are either \textbf{(1)}~not consistently {effective} in terms of ranking accuracy gains or \textbf{(2)}~impractically {inefficient} on larger datasets due to the combinatorial explosion problem of pairwise ranking with deep language models.
In this paper, we propose a novel tiered ranking architecture \cascader{} to maintain the ranking \emph{accuracy} of full ensembling while improving \emph{efficiency} considerably.
\cascader{} uses \lm{}s to %  reweight and 
rerank the outputs of more efficient base \kge{}s, relying on an adaptive subset selection scheme aimed at invoking the \lm{}s minimally while maximizing accuracy gain over the \kge. 
Extensive experiments demonstrate that \method{} improves MRR by up to 9 points over \kge{} baselines, setting new state-of-the-art performance on four benchmarks while improving efficiency by one or more orders of magnitude over competitive cross-modal baselines.
Our empirical analyses reveal that
diversity of models across modalities and
preservation of individual models' confidence signals
help explain the effectiveness of \cascader, and suggest promising directions for cross-modal cascaded architectures.
% Code and pretrained models
% will be made publicly available. 

% are available at 
% \url{https://github.com/tsafavi/cascader}. 
\end{abstract}

\section{Introduction}
\label{sec:intro}
Knowledge graphs (\kg{}s) are critical ingredients for applications across natural language processing, information retrieval, and biomedicine~\citep{weikum-etal-2020-machine}. 
Motivated by the observation that most \kg{}s have high precision but low coverage, %~\citep{galarraga-etal-2017-predicting-completeness}, 
the goal of \textbf{knowledge graph link prediction}, also known as \kg{} completion, is to automatically augment \kg{}s with new factual information by predicting missing links between  entities~\citep{nickel-etal-2015-relational-review}.

Link prediction is typically framed as a ranking problem in a multi-relational graph~\citep{bordes-etal-2013-translating}: 
Given a query consisting of a \emph{head} entity (e.g., \emph{aspirin}) and \emph{relation} type (e.g., \emph{treats}), the task is to rank candidate \emph{tail} entities by the likelihood that they answer the query and form a factual link the graph. 
Currently, the prevailing approach is to learn vector representations of entities and relations, or \textbf{knowledge graph embeddings} (\kge{}s), and use vector composition functions to score candidate links~\citep{ruffinelli-etal-2020-you}. While often effective at modeling structural \kg{} patterns~\citep{sun-etal-2019-rotate},
\kge{}s typically do not leverage textual information like entity descriptions in \kg{}s, even though such texts help ameliorate \kg{} sparsity and improve ranking accuracy~\citep{xie-etal-2016-entity-descriptions,chandak2022building}.

To address this gap, recent studies have proposed to \emph{ensemble} \kge{}s with advanced \textbf{language models} (\lm{}s) like BERT~\citep{devlin-etal-2019-bert} in order to integrate structure and text for link prediction~\citep{nadkarni-etal-2021-scientific,wang-etal-2021-structure-augmented}. 
These studies suggest the promise of cross-modal ensembles,
but results are inconclusive due to two challenges.
First, the cross-modal ensembles achieving the largest gains rely on impractically expensive models that require jointly encoding pairs of texts with deep \lm{}s~\citep{nadkarni-etal-2021-scientific}, greatly increasing the computational cost of inference---e.g., from a few minutes with a \kge{} to \emph{one month} with an \lm{} on the same dataset~\citep{kocijan-lukasiewicz-2021-knowledge}).
Second, the cross-modal ensembles that rely on more efficient ``Siamese'' dual-encoder \lm{} architectures do not necessarily improve performance over \kge{}s alone~\citep{wang-etal-2021-structure-augmented}. 

\begin{figure}
    \centering
    \includegraphics[width=0.35\textwidth]{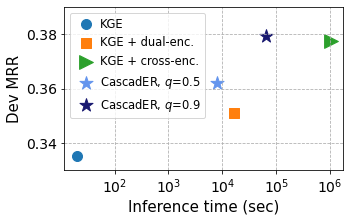}
    \caption{\cascader{} maintains accuracy (y-axis) while improving efficiency (x-axis) by 1+ orders of magnitude over our most competitive ensemble baseline on  \codex-M.
    % Dual-enc.: Dual-encoder \lm{}; Cross-enc: Cross-encoder \lm; see \S~\ref{prelim:single-modality} for details.
    We use a three-tier \cascader{} with dynamic pruning at quantiles $q=0.5$ and $q=0.9$ (\S~\ref{method:dynamic-pruning}). 
    }
    \label{fig:efficiency-base-comparison}
    \vspace{-.3cm}
\end{figure}

In this paper, we introduce a novel cross-modal ensemble that simultaneously maximizes ranking \emph{accuracy}, i.e., mean reciprocal rank (MRR) and hits@$k$, while maintaining \emph{efficiency}. 
Our approach, \textbf{\cascader}, is a tiered ranking architecture that uses a sequence of \lm{}s of increasing complexity to adaptively reweight and rerank the outputs of more efficient base \kge{}s. 
\cascader{} relies on a novel subset selection scheme:
At each tier $t$ of the architecture, we predict the minimal set of candidates that should progress to be reweighted at tier $t+1$.
By passing progressively smaller sets of outputs from one tier to the next, \cascader{} balances accuracy gain and efficiency, as it minimizes invocation of the more computationally complex \lm{}s further down the cascade. 

Evaluated on five link prediction datasets, \cascader{} achieves state-of-the-art performance, improving MRR over the strongest \kge{} baseline by up to 9 points, while also improving efficiency by orders of magnitude over our most accurate but computationally intensive ensemble baseline  (Figure~\ref{fig:efficiency-base-comparison}). 
Our qualitative analyses illuminate how cross-modal ensembling uniquely exploits complementary signals among graph and text models: We observe that promoting diversity and preserving confidence signals among the models in the ensemble help explain \cascader{}'s excellent performance, suggesting promising directions for research in cross-modal cascaded architectures. 

\section{Preliminaries}
\label{sec:prelim}
In this section, we provide preliminaries on single-modality and cross-modal approaches for link prediction. 
We discuss additional related work in Appendix~\ref{appendix:related-work}. 

\subsection{Problem definition}
\label{prelim:setup}

We consider the task of \textbf{ranking-based link prediction} in a knowledge graph $\G$ consisting of entities $\V$, relations $\relations$, and factual (\emph{head}, \emph{relation}, \emph{tail}) triples $(h, r, t)  \in \V{} \times \relations{} \times \V$.
The link prediction task consists of two directionalities:
Score all tail entities $\hat{t} \in \V$ to ``answer queries''---that is, complete known \kg{} links---$(h, r, ?)$, and score all head entities $\hat{h} \in \V$ to answer queries  $(?, r, t)$. 
The evaluation metrics are mean reciprocal rank (MRR), or the average reciprocal of each gold answer entity's rank over all queries, and hits@$k$, or the proportion of queries for which the gold answer entity is ranked in the top-$k$ candidates. 

A link prediction model is a scoring function 
$f : \V{} \times \relations{} \times \V{} \rightarrow \mathbb{R}$ that outputs real values indicating the plausibility of factual links in $\G$. 
At inference time, assume that we have $\numTest$ link prediction queries.
% of interest, half of type $(h, r, ?)$ and half of type $(?, r, t)$.
For each query we have $|\V|$ potential answers, which are the entities in the \kg{}. 
We define a query-answer \textbf{score matrix} $\scoreMatrix \in \mathbb{R}^{\numTest \times |\V|}$, 
in which $S_{ij}$ denotes the predicted probability that entity $j$ answers link prediction query $i$ to form a factual link in $\G$. 
Then, for each query $i$, all candidate entities $j$ are ranked by their score descending, and the model's ranking accuracy is evaluated using these rankings. 

\subsection{Single-modality link prediction}
\label{prelim:single-modality}

% We provide a brief overview of single-modality link prediction approaches.
% See Appendix~\ref{appendix:prelim-single-modality} for additional details. 

\paragraph{Structure-based}
The most competitive structure-only approaches to link prediction are shallow knowledge graph embeddings (\textbf{\kge}s), which are decoder models that train entity and relation embeddings by optimizing with ranking losses. 
The main architectural distinction among different \kge{}s is how embeddings are combined to produce ranking scores for links, as both additive~\citep{bordes-etal-2013-translating,sun-etal-2019-rotate} and multiplicative~\citep{yang-etal-2014-distmult,trouillon-etal-2016-complex,balazevic-etal-2019-tucker} scoring functions have been proposed. 
For details, we refer the reader to relevant surveys~\citep{wang-etal-2017-kge-survey,ji-etal-2020-survey}. 

\paragraph{Text-based}
Recent text-based approaches to link prediction rely on advanced encoder language models (\lm{}s) like BERT~\citep{devlin-etal-2019-bert} based on the Transformer architecture~\citep{vaswani-etal-2017-attention}. 
Let $X_h = [w_1, \hdots, w_{h}]$ denote the description of head entity $h$ 
and $X_t = [w_1, \hdots, w_{t}]$ the description of tail entity $t$;
for example, given the entity \emph{aspirin}, a corresponding description could be ``aspirin is known as a salicylate and a nonsteroidal anti-inflammatory drug.'' 
Siamese or \textbf{dual-encoder} link prediction approaches assume two \lm{}s, potentially with shared weights~\citep{wang-etal-2021-structure-augmented}. 
As input, one \lm{} takes the head entity description [\CLS{}, $X_h$, \SEP{}] and the other takes the tail entity description [\CLS{}, $X_t$, \SEP{}],
where \CLS{} and \SEP{} refer to the \lm{}'s special classification and delimiter tokens. 
Both encoders output embeddings of their inputs, which are optimized such that linked entity pairs in the \kg{} are scored highly compared to negative samples.\footnote{Previous studies have found that the relation text may be omitted, and that including a relation disambiguation loss term in training is sufficient~\citep{kim-etal-2020-multi,nadkarni-etal-2021-scientific}.}

% is usually more important than encoding the relation text for performance~\citep{kim-etal-2020-multi,nadkarni-etal-2021-scientific}.}

In contrast to dual-encoders, \textbf{cross-encoder} \lm{} approaches pack entity description pairs into a single sequence [\CLS{}, $X_h$, \SEP{}, $X_t$, \SEP{}], and perform full cross-attention over all tokens in the sequence~\citep{yao-etal-2019-kg-bert,nadkarni-etal-2021-scientific}.
At the output of the encoder, these approaches stack a scoring layer and train with ranking loss~\citep{kim-etal-2020-multi}.  
Cross-encoder \lm{}s are typically more powerful for pairwise text ranking than dual-encoders~\citep{luan-etal-2021-sparse}. 
However, they are less efficient.  
Whereas dual-encoders can precompute all text embeddings and score text pairs at test time with fast vector dot products~\citep{karpukhin-etal-2020-dense}, 
cross-encoders must jointly encode and score each text pair, which is impractically slow for large-scale text ranking~\citep{reimers-gurevych-2019-sentence}. 

\subsection{Cross-modal link prediction}
\label{prelim:ensemble}

Structure and text have recently been integrated for link prediction by ensembling \kge{}s and \lm{}s with additive reweighting~\citep{wang-etal-2021-structure-augmented,nadkarni-etal-2021-scientific}. 
Given a query $i$ and candidate answer $j$, additive reweighting outputs a new link prediction score as the convex combination of the base models' scores: 
\begin{align}
    \label{eq:additive-reweighting}
S_{ij}^{\textrm{ens}} = \alpha \cdot S_{ij}^{\textrm{\kge}} + 
\left(1 - \alpha\right) \cdot S_{ij}^{\textrm{\lm}},
\end{align}
where the weight $\alpha \in [0, 1]$ is a hyperparameter tuned on a held-out set.

% \begin{table}[!t]
%     \caption{
%     Additive ensembling improves MRR over the base \kge{} by up to 4 points dev MRR.
%     }
%     \label{table:ensemble-comparison}
%     \centering
%     \resizebox{0.55\textwidth}{!}{
%         \begin{tabular}{l ll}
%         \toprule
%         & \codex-M dev & \repodb{} dev \\ 
%         \toprule 
%         State-of-the-art \kge{} & 0.3351 & 0.4899 \\ 
%         % Dual-encoder \lm & 0.2727 & 0.3405 \\ 
%         % Cross-encoder \lm & 0.2032 & 0.2688 \\ 
%         % \midrule 
%         \kge{} + dual-encoder \lm & 0.3509 (+1.58) & 0.5003 (+1.04) \\
%         \kge{} + cross-encoder \lm & 0.3777 (+4.26) & 0.5209 (+3.10) \\
%         \bottomrule
%         \end{tabular}
%     }
% \end{table}

As shown in Figure~\ref{fig:efficiency-base-comparison}, additive reweighting can significantly improve link prediction ranking accuracy, up to 4 points MRR.
% by combining a \kge{} with a cross-encoder \lm{}. 
However, Figure~\ref{fig:efficiency-base-comparison} also shows that ensembling increases inference complexity.  % by orders of magnitude: 
Whereas the base \kge{} requires under one minute to score all query-answer pairs on a single NVIDIA Quadro RTX 8000 GPU, the \kge{} + dual-encoder ensemble takes around 3 hours, and the \kge{} + cross-encoder ensemble takes over \emph{11 days} on the same hardware.
This added expense is due to the fact that deep Transformer \lm{}s typically consist of 12+ layers, and the complexity of Transformer encoding scales quadratically with the input length. 
Moreover, as discussed previously, cross-encoders must jointly encode all query/answer pairs, which further increases their computational cost.

\section{Methodology}
\label{sec:method}
Assuming we are willing to pay some computational cost to improve link prediction performance, how can we achieve the \emph{accuracy} of the cross-encoder ensemble while maintaining the \emph{efficiency} of the dual-encoder ensemble, as shown in Figure~\ref{fig:efficiency-base-comparison}? 
Our answer is \textbf{\cascader{}}, a cross-modal cascade architecture that achieves a delicate balance between these goals. 

\subsection{\cascader{} overview}
\label{method:cascader}

\begin{figure}
    \centering
    \includegraphics[width=0.9\textwidth]{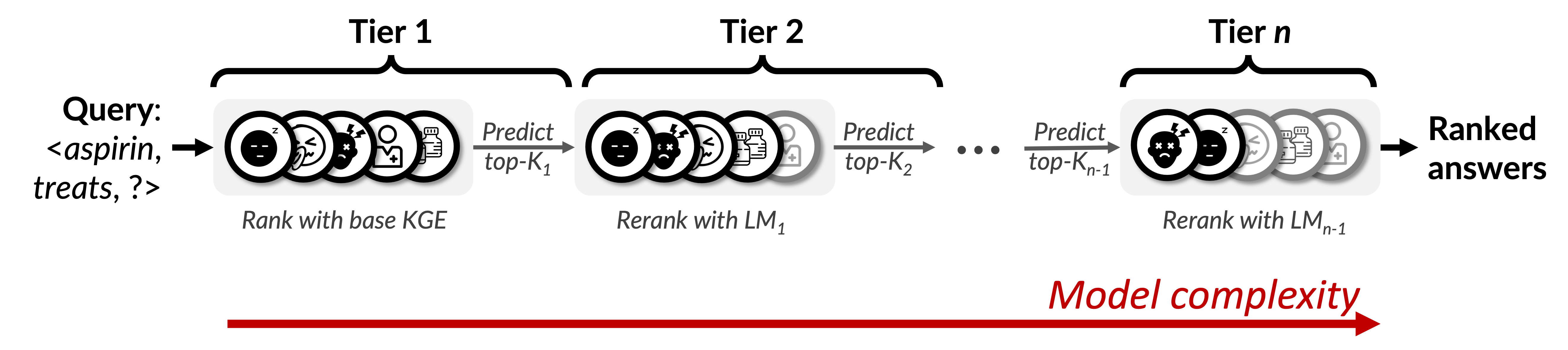}
    \caption{\cascader{} sequential reranking architecture. 
    }
    \label{fig:schematic}
    % \vspace{-.5cm}
\end{figure}

As illustrated in Figure~\ref{fig:schematic}, \cascader{} is a progressive reranking architecture. 
We first obtain a base set of link prediction scores with an efficient \kge{}, then use increasingly complex \lm{}s to rerank the base scores on progressively smaller sets of outputs---specifically, only the highly-ranked, most promising candidates, while leaving the rankings of the less promising candidates static.  
\cascader{} thus benefits from the performance gains of cross-modal ensembles without the full computational overhead (c.f. Figure~\ref{fig:efficiency-base-comparison}).

As input, we are given $n \geq 2$ trained link prediction models $\{f^{(i)}, \, i = 1 \hdots n\}$ consisting of one \kge{} and one or more \lm{}s. 
We sort the models by computational complexity, leading to an ordered sequence $(f^{(1)}, \hdots, f^{(n)})$ in which $f^{(1)}$ is the \kge{} and the subsequent models are \lm{}s in ascending order of complexity (i.e., dual-encoders before cross-encoders). 
We first use the \kge{} to score all query/answer pairs in the inference set, leading to a score matrix $\scoreMatrix^{(1)} \in \realNumbers^{\numTest \times |\V|}$
in which $S_{ij}^{(1)}$ denotes the \kge{}'s plausibility score between query $i$ and entity $j$. 
Then, at each tier $t = 1, \hdots, n - 1$, we apply a pruning function that, for each query $i$, selects a subset of candidate answer entities $\V_i^{(t)} \subseteq \V$ to be reranked by the next-tier \lm{} $\secondRanker$; we postpone the discussion of pruning strategies to \S~\ref{method:static-pruning} and~\ref{method:dynamic-pruning}. 

For link prediction query $i$ (that is, a query entity and relation type) and candidate answer entity $j$, we define the additive reranking function between tiers $t$ and $t + 1$ as follows:
\begin{align}
    \label{eq:reranking}
\mathbbm{I}_{j \in \V_{i}^{(t)}} \left[  
\alpha^{(t)} \cdot S_{ij}^{(t)} + 
\left(1 - \alpha^{(t)}\right) \cdot S_{ij}^{(t+1)}
\right] 
+ 
\mathbbm{I}_{j \not\in \V_{i}^{(t)}} \left[ S_{ij}^{(t)} \right], 
\end{align}
in which $\mathbbm{I}$ denotes the set indicator function, $S_{ij}^{(t)}$ denotes the query-answer score output at tier $t$, and $\alpha^{(t)} \in [0, 1]$ is a hyperparameter that controls the additive influence of model $\firstRanker$ in reranking the candidates in $\V_i^{(t)}$.
Intuitively, \eqref{eq:reranking} states that 
if the candidate answer entity $j$ is within the subset $\V_i^{(t)}$ of candidates progressed to tier $t+1$, then we reweight its score with that of  model $f^{(t+1)}$ at tier $t+1$, else we do not reweight its score.
% If the candidate answer entity $j \not\in \V_i^{(t)}$, then we do not reweight its score at tier $t+1$. 

\subsection{Static candidate pruning}
\label{method:static-pruning}

Candidate pruning with \cascader{} should balance two aims:
\textbf{(1)}~\emph{Coverage}: For each link prediction query $(h, r, ?)$ or $(?, r, t)$, progress a sufficiently large set of candidate entities to the following tier, in order to increase coverage of the correct candidates; and 
\textbf{(2)}~\emph{Efficiency}: Progress as few candidates as possible to avoid unnecessarily invoking the next-tier reranking model on candidates that would not benefit from it.

A straightforward pruning approach used in information retrieval cascades is to progress only the top-$k$ candidates from tier to tier using a predefined, global value of $k$~\citep{wang-etal-2011-cascade,matsubara-etal-2020-reranking}.
This value helps control the accuracy-efficiency tradeoff, as 
a smaller $k$ decreases coverage of promising candidates but improves efficiency, whereas a larger $k$ increases coverage of promising candidates but incurs more computational cost. 
Formally, given query $i$ and a selected value of $k$, we define \textbf{static pruning} as selecting the subset of candidates $\V_i^{(t)}$ such that
$
\V_i^{(t)} = \argmax_{\V_i^{(t)} \subset \V \textrm{ and } |\V_i^{(t)}| = k} \sum_{j=1}^{|\V|} S_{ij}^{(t)}. 
$

The key challenge with static pruning  is selecting the ``right'' value of $k$.
One solution is to set $k$ ad-hoc, e.g., $k=100$~\citep{matsubara-etal-2020-reranking}. 
However, this approach may result in suboptimal performance from the accuracy or efficiency perspectives. 
To address this challenge, we propose a more principled strategy that searches for the best value of $k$ per dataset and per tier $t$ of the cascade. %  in terms of MRR on a held-out development set. 
Given tier $t$ and held-out query $i$, we obtain the cascade's rank $R_i^{(t)}$ of the gold answer entity. 
We construct a distribution of ranks $R_i^{(t)}$ over all hold-out queries, and use quantiles of this distribution to choose the grid of $k^{(t)}$ over which to search. % , using dev MRR as our evaluation metric. 
For example, quantiles of 0.5, 0.75, and 0.9 means that we search over $k^{(t)}$ equal to the median, 75th percentile, and 90th percentile of ranks $R_i^{(t)}$, which ensures that our selected $k^{(t)}$ is tailored to each dataset and each tier of reranking. 

\subsection{Dynamic candidate pruning}
\label{method:dynamic-pruning}

We propose to extend static pruning to an adaptive \textbf{dynamic pruning} approach.
At any tier $t$ of \cascader, we improve the quality of the current ranking by \emph{predicting} how many top-ranked candidates \emph{for each query} we should progress to the next tier. 
That is, given a query $i$ and tier $t$, we predict an integer $\hat{k}_i^{(t)}$ that represents the number of candidates to progress to tier $t + 1$. 
To achieve this, we train a lightweight model to predict the rank of the gold answer entity using quantile regression~\citep{koenker-hallock-2001-quantile-regression}. 
Formally, given a quantile $q$ and the rank of the gold answer entity $R_i^{(t)}$ at tier $t$, we train a regressor to predict $\hat{k}_i^{(t)}$ by minimizing
$
\loss_{q}(\hat{k}_{i}^{(t)}, R_i^{(t)}) = \max \left[ q(R_i^{(t)} - \hat{k}_{i}^{(t)}), (q-1)(R_i^{(t)} - \hat{k}_{i}^{(t)}) \right]. 
$
As input features, we represent the $i$-th query by its sorted $|\V|$-dimensional score distribution $S_{i1}^{(t)}, \hdots S_{i|\V|}^{(t)}$ from tier $t$ of the cascade, hypothesizing that these score distributions encode uncertainty information correlated to the  relative ``difficulty'' of queries. 
In practice, we implement our regressor as a single-layer MLP trained on a random half of the dev set, and validated on the remaining dev examples.\footnote{Note that if the dev set is unavailable, we can alternatively hold out a small subset of train examples.}
We will subsequently show that this approach boosts \cascader{}'s ability to balance accuracy and efficiency compared to static pruning.

\section{Evaluation}
\label{sec:eval}
% In this section, we evaluate \cascader{} on five link prediction datasets. 
As introduced in \S~\ref{prelim:setup}, we evaluate \cascader{} for the link prediction task. 
Our evaluation metrics are MRR and hits@$k$ for $k \in \{1, 3, 10\}$. 
Following the literature standard~\citep{ruffinelli-etal-2020-you}, we report metrics in the filtered setting to avoid false negatives. 

% masking out all known answers to test queries other than the gold answer entity in question, in order to avoid false negatives.  

\begin{table}[t!]
    \caption{
    Statistics of the existing \kg{} link prediction datasets considered in our experiments.
    Avg. desc. length refers to the average description of an entity in the \kg. 
    }
    \label{table:datasets}
    \centering
    \resizebox{0.6\textwidth}{!}{
        \begin{tabular}{l c c c c c c c c}
        \toprule
        && \multicolumn{5}{c}{Structure} && \multirow{2}{*}{Avg. desc. length} \\ 
        \cline{3-7}
        && $|\V|$ & $|\relations|$ & \# train & \# dev & \# test &&  \\
        \toprule
        \codexS{} && 2,034 & 42 & 32,888 & 1827 & 1828 
        % & 0.01767 
        && 259.24 \\ 
        \repodb{} && 2,748 & 1 & 5,342 & 667 & 668 
        % & 0.00176 
        && 55.46 \\ 
        \fb{} && 14,541 & 237 & 272,115 & 17,535 & 20,466 
        % & 0.00293 
        && 138.95  \\ 
        \codexM{} && 17,050 & 51 & 185,584 & 10,310 & 10,311
        % & 0.00142 
        && 159.48 \\ 
        \wn{} && 40,943 & 11 & 86,835 & 3,034 & 3,134 
        % & 0.00011 
        && 13.91 \\ 
        \bottomrule
        \end{tabular}
    }
\end{table}

\subsection{Datasets}
\label{exp:data}

We consider the following link prediction benchmarks, as shown in Table~\ref{table:datasets}:  
\textbf{\fb{}}~\citep{toutanova-chen-2015-observed},
\textbf{\wn{}}~\citep{dettmers-etal-2018-convolutional}, 
\textbf{\codexS{}} and \textbf{\codexM{}}~\citep{safavi-koutra-2020-codex}, 
and
\textbf{\repodb{}}~\citep{brown-patel-2017-standard,nadkarni-etal-2021-scientific}. 
In terms of content, \fb{}, \codexS{}, and \codexM{} comprise encyclopedic knowledge drawn from Freebase and Wikidata,
\wn{} is a subset of the WordNet semantic network, and \repodb{} is a subset of the RepoDB drug repurposing biomedical database~\citep{brown-patel-2017-standard}. 
% Table~\ref{table:datasets} in Appendix~\ref{appendix:data} provides structural and textual statistics for the datasets. 
For all datasets, we use the standard splits provided by the authors. 
We use the entity descriptions provided by~\citet{wang-etal-2021-structure-augmented} for \fb{} and \wn{},
 \citet{nadkarni-etal-2021-scientific} for \repodb{},
and \citet{safavi-koutra-2020-codex} for \codexS{} and \codexM{}.

\subsection{Baselines}
\label{exp:baselines}

% We consider the following baselines:  

\paragraph{\kge{} baselines}
We consider the competitive
\textbf{RESCAL}~\citep{nickel-etal-2011-three}, \textbf{TransE}~\citep{bordes-etal-2013-translating}, 
\textbf{ComplEx}~\citep{trouillon-etal-2016-complex}, 
and \textbf{RotatE}~\citep{sun-etal-2019-rotate} \kge{}s. 

\paragraph{\lm{} baselines}
We consider the \textbf{StAR dual-encoder} architecture~\citep{wang-etal-2021-structure-augmented} and the \textbf{KG-BERT cross-encoder} architecture~\citep{yao-etal-2019-kg-bert}, both trained in a multi-task setting with triple classification, margin ranking, and relation classification losses following the literature~\citep{kim-etal-2020-multi}. 
Due to the inference cost of KG-BERT on the larger datasets \fb{} and \wn{} (e.g., around one month for \fb~\citep{kocijan-lukasiewicz-2021-knowledge}), 
we report performance for these datasets from~\citep{kim-etal-2020-multi}. 

\paragraph{Ensemble baselines}
We consider the following additive ensembling baselines as defined in \S~\ref{prelim:ensemble}: 
%controlled by a weighting hyperparameter $\alpha$ tuned on the dev set:
\textbf{\kge{} + \kge{}} ensembles the two strongest \kge{} baselines in terms of dev MRR; %  on the validation set; 
\textbf{\kge{} + StAR}~\citep{wang-etal-2021-structure-augmented} ensembles the best \kge{} with StAR;
and \textbf{\kge{} + KG-BERT}~\citep{nadkarni-etal-2021-scientific} ensembles the best \kge{} with KG-BERT.

\paragraph{SOTA}
We report the best published performance of which we are aware as of April 2022:
NBFNet~\citep{zhu-etal-2021-nbfnet} on \fb{},
self-adaptive \kge{} + \lm{} ensembling~\citep{wang-etal-2021-structure-augmented} on \wn{},
and ComplEx for the CoDEx datasets.
Note that \repodb{} has not been considered under the full entity ranking setting before, as~\citet{nadkarni-etal-2021-scientific} used a partial sample of negatives for each test query, so we do not report SOTA for \repodb.

\subsection{\cascader{}} 
\label{eval:cascader}

Our first-tier \kge{} in \cascader{} is the best-performing baseline \kge{} in terms of dev MRR. 
We search for the optimal cascade in terms of dev MRR among the following hyperparameters:
The choice of \lm{}s (StAR dual-encoder, KG-BERT cross-encoder, or both);
candidate pruning strategy (static versus dynamic);
quantile $q \in \{0.5, 0.75, 0.9, 0.95\}$; 
and weighting hyperparameter $\alpha^{(t)} \in [0.05, 0.95]$ at each tier. 
Appendix~\ref{appendix:model-selection} provides additional details on hardware, software, and model selection.

\subsection{Inference budget}
\label{eval:budget}
We imposed an inference time limit on \cascader{} and all baselines.
First, we set a global maximum inference time of 24 hours on a single NVIDIA Quadro RTX 8000 GPU with 48 GB of RAM as a hard limit. 
Then, we set dataset-specific budgets $\leq$ 24H roughly as a function of the test set size and \kg{} size. 
For example, \wn{} has fewer test queries and shorter text lengths (Table~\ref{table:datasets}) than \fb, but 3$\times$ as many answer entities. 
Given a budget of 24H for \fb, our largest dataset, we decided to allow 25\% of that time to \wn{} to account for these differences.
Concretely, 
we set inference time limits to 2 hours for our two smallest datasets \codexS{} and \repodb{}, 24 hours for our two largest datasets \codexM{} and \fb, 
and 6 hours for \wn{}.
To understand the impact of our chosen time limits, we vary these budgets in an ablation in \S~\ref{eval:results}. 

\begin{table}[!t]
    \caption{\cascader{} outperforms or is competitive with the state of the art on \fb{} and \wn. 
    \textbf{\underline{Bold + underline}}: Best performance.
    \underline{Underline}: Second-best performance. 
    The performance of StAR, KG-BERT, and SOTA are reported from papers referenced in \ref{exp:baselines}. 
    OOT refers to out-of-time using our inference cost budget (\S~\ref{eval:budget}). 
    }
    \label{table:fb-wn-results}
    \centering
    \resizebox{0.7\textwidth}{!}{
        \begin{tabular}{l cccc c cccc}
        \toprule
        &
        \multicolumn{4}{c}{\textbf{\fb}} &&
        \multicolumn{4}{c}{\textbf{\wn}} \\
        % \multicolumn{4}{c}{\textbf{\repodb}} \\
        \cline{2-5}
        \cline{7-10}
        &
            MRR & H@1 & H@3 & H@10 &&
            MRR & H@1 & H@3 & H@10 \\ 
        \midrule 
        RESCAL & 0.3559 & 0.2629 & 0.3926 & 0.5406 && 0.4666 & 0.4387 & 0.4797 & 0.5172 \\
        TransE & 0.3128 & 0.2206 & 0.3473 & 0.4973 && 0.2278 & 0.0531 & 0.3682 & 0.5201 \\ 
        ComplEx & 0.3477 & 0.2533 & 0.3836 & 0.5359 && 0.4749 & 0.4381 & 0.4898 & 0.5474 \\
        RotatE & 0.3333 & 0.2396 & 0.3676 & 0.5218 && 0.4781 & 0.4395 & 0.4941 & 0.5527 \\
        \midrule 
        StAR & 0.296 & 0.205 & 0.322 & 0.482 && 0.401 & 0.243 & 0.491 & 0.709 \\
        KG-BERT & 0.267 & 0.172 & 0.298 & 0.458 && 0.331 & 0.203 & 0.383 & 0.597 \\
        \midrule 
        \kge{} + \kge{} & 0.3630 & 0.2672 & 0.4016 & 0.5535 && 0.4900 & 0.4521 & 0.5016 & 0.5617 \\
        \kge{} + StAR & 0.3643 & 0.2709 & 0.3989 & 0.5522 && 0.5385 & 0.4716 & 0.5645 & 0.6651 \\
        \kge{} + KG-BERT & OOT & OOT & OOT & OOT && OOT & OOT & OOT & OOT \\
        \midrule 
        SOTA & \textbf{\underline{0.415}} & \textbf{\underline{0.321}} & \textbf{\underline{0.454}} & \textbf{\underline{0.599}} && \underline{0.551} & \underline{0.459} & \underline{0.594} & \underline{0.732} \\ 
        \midrule 
        \textbf{\cascader{}} & \underline{0.3860} & \underline{0.2903} & \underline{0.4231} & \underline{0.5782} && 
        \textbf{\underline{0.5651}} & \textbf{\underline{0.4756}} & \textbf{\underline{0.6126}} & \textbf{\underline{0.7379}} \\ 
        \bottomrule
        \end{tabular}
    }
\end{table}

% \afterpage{
 \begin{table}[!t]
    \caption{
    \cascader{} achieves state-of-the-art test performance on \codexM{} and \repodb. 
    % \textbf{\underline{Bold + underline}}: Best performance.
    % \underline{Underline}: Second-best performance. 
    % OOM refers to out-of-memory during training.
    % OOT refers to out-of-time using our inference cost budget. 
    }
    \label{table:codexm-repodb-results}
    \centering
    \resizebox{0.7\textwidth}{!}{
        \begin{tabular}{l cccc c cccc}
        \toprule
        & 
        \multicolumn{4}{c}{\textbf{\codex-M}} &&
        \multicolumn{4}{c}{\textbf{\repodb}} \\
        \cline{2-5}
        \cline{7-10}
        & 
        MRR & H@1 & H@3 & H@10 
        &&
        MRR & H@1 & H@3 & H@10  \\ 
        \midrule 
        RESCAL & 
            0.3173 & 0.2444 & 0.3477 & 0.4557
            &&
            0.4351 & 0.3144 & 0.4903 & 0.6767 \\ 
        TransE & 
            0.3026 & 0.2232 & 0.3363 & 0.4535
            &&
            0.3472 & 0.2043 & 0.3728 & 0.6400 \\
        ComplEx & 
            0.3365 & 0.2624 & 0.3701 & 0.4758
            &&
            0.4620 & 0.3406 & 0.5225 & 0.7043 \\ 
        RotatE & 
            OOM & OOM & OOM & OOM
            &&
            0.2971 & 0.1811 & 0.4903 & 0.5314 \\ 
        \midrule 
        StAR & 
            0.2726 & 0.1888 & 0.3042 & 0.4342
            &&
            0.3472 & 0.2043 & 0.4102 & 0.6400 \\ 
        KG-BERT & 
            OOT & OOT & OOT & OOT
            &&
            0.2991 & 0.1602 & 0.3428 & 0.5996 \\ 
        \midrule 
        \kge{} + \kge{} & 
            0.3466 & 0.2695 & 0.3808 & 0.4925
            &&
            0.4637 & 0.3398 & 0.5262 & 0.7081 \\ 
        \kge{} + StAR &
            \underline{0.3554} & \underline{0.2767} & \underline{0.3901} & \underline{0.5064} 
            &&
            0.4774 & 0.3496 & 0.5434 & 0.7208 \\ 
        \kge{} + KG-BERT & 
            OOT & OOT & OOT & OOT
            &&
            \underline{0.5101} & \underline{0.3713} & \underline{0.5771} & \underline{0.7799} \\ 
        \midrule 
        SOTA &  
            0.3365 & 0.2624 & 0.3701 & 0.4758
            &&
            - & - & - & - \\ 
        \midrule 
        \textbf{\cascader{}} &  
            \textbf{\underline{0.3830}} & \textbf{\underline{0.2998}} & \textbf{\underline{0.4221}} & \textbf{\underline{0.5423}}
            &&
            \textbf{\underline{0.5156}} & \textbf{\underline{0.3817}} & \textbf{\underline{0.5831}} & \textbf{\underline{0.7814}} \\ 
        \bottomrule
        \end{tabular}
    }
\end{table}

% \begin{table}[!t]
%     \caption{\cascader{} sets a new state of the art on the drug repurposing benchmark \repodb{}. 
%     }
%     \label{table:repodb-results}
%     \centering
%     % \def\arraystretch{1.1}
%     \resizebox{0.5\textwidth}{!}{
%         \begin{tabular}{l cccc}
%         \toprule
%         & MRR & H@1 & H@3 & H@10 \\ 
%         \midrule 
%         RESCAL & 0.4351 & 0.3144 & 0.4903 & 0.6767 \\ 
%         TransE & 0.3472 & 0.2043 & 0.3728 & 0.6400 \\
%         ComplEx & 0.4620 & 0.3406 & 0.5225 & 0.7043 \\ 
%         RotatE & 0.2971 & 0.1811 & 0.4903 & 0.5314 \\ 
%         \midrule 
%         StAR & 0.3472 & 0.2043 & 0.4102 & 0.6400 \\
%         KG-BERT & 0.2991 & 0.1602 & 0.3428 & 0.5996 \\
%         \midrule 
%         \kge{} + \kge{} & 0.4637 & 0.3398 & 0.5262 & 0.7081 \\ 
%         \kge{} + StAR & 0.4774 & 0.3496 & 0.5434 & 0.7208 \\ 
%         \kge{} + KG-BERT & \underline{0.5101} & \underline{0.3713} & \underline{0.5771} & \underline{0.7799} \\ 
%         \midrule 
%         \textbf{\cascader{}} & \textbf{\underline{0.5156}} & \textbf{\underline{0.3817}} & \textbf{\underline{0.5831}} & \textbf{\underline{0.7814}} \\ 
%         \bottomrule
%         \end{tabular}
%     }
% \end{table}

\subsection{Results and discussion}
\label{eval:results}

Table~\ref{table:fb-wn-results} and \ref{table:codexm-repodb-results}
provide link prediction performance results for \fb{} and \wn{}, and \codexM{} and \repodb{}, respectively;
Table~\ref{table:codexs-results} in Appendix~\ref{appendix:additional-results} provides results on \codexS{}, omitted here for brevity. 
% and~\ref{table:repodb-results} provide link prediction performance results for \fb{} and \wn{}, the \codex{} datasets, and \repodb, respectively.
We observe that \cascader{} achieves robust and appreciable gains over baselines across datasets, setting a new state of the art on \wn{}, \codexS{}, \codexM{}, and \repodb{}, and performing second to the reported SOTA on \fb. 
It outperforms the best \kge{} by up to 8.70 points MRR (\wn) and the best \lm{} by up to 16.84 points MRR (\repodb), demonstrating that cross-modal ensembling can significantly improve upon single-modality approaches.

We also remark that full additive ensembling is \emph{not} necessary to maximize accuracy. 
Our \kge{} + \kg-BERT additive ensemble baseline is competitive on  \codexS{} and \repodb{}, but it encounters out-of-time errors on  the other datasets. 
By contrast, \cascader{} achieves state-of-the-art or competitive performance on all five datasets while staying within our time limits. 
This suggests that full additive ensembling is not necessary to achieve the majority of gains in link prediction, and that cascaded reranking is sufficient. 

\begin{figure}[!t]
    \centering
    \begin{subfigure}{0.35\columnwidth}
        \includegraphics[width=\textwidth]{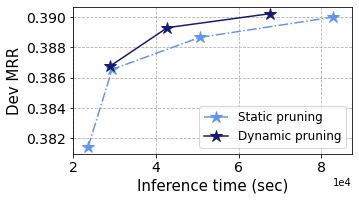}
        \caption{
            Dynamic pruning balances accuracy and efficiency better than static pruning.
            % We consider a three-tier cascade with pruning only between tiers two and three.
        }
        \label{fig:pareto-pruning}
    \end{subfigure}
    ~
    \begin{subfigure}{0.35\columnwidth}
        \includegraphics[width=\textwidth]{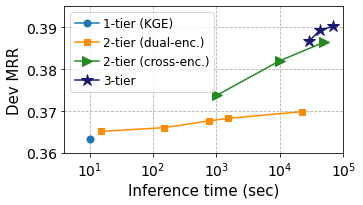}
        \caption{
            \cascader{} is effective at any level of cost constraint compared to the base \kge. 
            % For the two-tier cascades, we use dynamic pruning between tiers one and two. 
            % For the three-tier cascade, we use dynamic pruning only between tiers two and three.
        }
        \label{fig:pareto-num-tiers}
    \end{subfigure}
    \caption{
        Top-left corner is best: Pareto curve analysis on the dev set of \fb. 
        We use quantiles $q \in \{0.5, 0.75, 0.9, 0.95, 1\}$ in our analyses and exclude any quantiles that lead to \cascader{} exceeding our inference time limit of 24 hours. 
    }
    \label{fig:pareto}
\end{figure}

\paragraph{Pareto curve analysis}
We provide a Pareto curve analysis to characterize the accuracy-efficiency tradeoff of \cascader{}. 
In Figure~\ref{fig:pareto}, we plot the accuracy (dev MRR) and efficiency (inference cost in wall-clock time) against \cascader{}'s key hyperparameters, the candidate pruning strategy and the number of tiers. 
We observe that dynamic pruning balances accuracy and efficiency better than static pruning.
Figure~\ref{fig:pareto-pruning} shows that dynamic pruning leads to steeper MRR improvements than static pruning, with comparable inference times. 

Consistent with the information retrieval literature~\citep{matsubara-etal-2020-reranking,luan-etal-2021-sparse}, cross-encoders improve ranking performance more than their dual-encoder counterparts. 
Figure~\ref{fig:pareto-num-tiers} confirms that two-tiered \cascader{} with a cross-encoder is much more accurate than two-tiered \cascader{}  with a dual-encoder. 
At an inference time of around 1000 seconds, our two-tiered dual-encoder and cross-encoder architectures achieve 0.3683 and 0.3738 MRR respectively, suggesting that reranking \emph{very few} candidates with a cross-encoder is often more beneficial than reranking \emph{many} candidates with a dual-encoder. 

\paragraph{Parameter comparison}
A natural question is whether \cascader{}'s improvement is simply a result of increased representational capacity, i.e., more parameters. 
Depending on the dataset, the number of \kge{} parameters is on the order of 100K-10M, scaling linearly (for most architectures) with the number of entities and relations in the \kg. 
By contrast, each \lm{}, when implemented with BERT-Base, has roughly 100M parameters, regardless of \kg{} size; therefore, \cascader{} has more parameters than all \kge{} baselines. 
However, note that \lm{}s \emph{alone} for the link prediction task are not competitive with \kge{}s, as shown in Tables~\ref{table:fb-wn-results} and~\ref{table:codexm-repodb-results}, which suggests that increased parameter count  does not fully explain performance improvement.
Moreover, the Pareto improvements we observe are in comparison to ensembles of the same number of parameters (e.g., 2-tier dual-encoding \cascader{} vs 2-tier cross-encoding \cascader{}, Figure~\ref{fig:pareto-num-tiers}), where we achieve up to an order of magnitude efficiency improvement at the same or better accuracy.

 \begin{table}[!t]
    \caption{
    \cascader{} still maintains SOTA performance (MRR) under tightened inference budgets: Inference time limit ablation.
    ``Full budget'' refers to the time limits imposed in \S~\ref{eval:budget}, and ``half budget'' refers to a time limit of half that. 
    }
    \label{table:budget-ablation}
    \centering
    \resizebox{0.5\textwidth}{!}{
        \begin{tabular}{l c c}
        \toprule
        & \codexM{} & \wn{} \\ 
        \toprule
        Best baseline & 0.3554 & 0.5510 \\
        \cascader{}, half budget & 0.3787 & 0.5590 \\
        \cascader{}, full budget & 0.3830 & 0.5651 \\ 
        \bottomrule
        \end{tabular}
    }
\end{table}

\paragraph{Inference budget ablation}
We next analyze the impact of changing the inference budget (\S~\ref{eval:budget}) on \cascader{}'s performance.
As shown in Table~\ref{table:budget-ablation}, \cascader{}’s performance does not change drastically when tightening the budget. 
Halving the inference budget for \codexM{} from 24H to 12H, we get a test accuracy of 0.3787 MRR for \cascader{}, a drop of < 1 point MRR from our best \cascader{} under 24H, and still an improvement of 2 points MRR over the next-best baseline. 
Similarly, halving the inference budget for \wn{} from 6H to 3H, we again observe a drop of < 1 point MRR.

\paragraph{Qualitative analysis}
Finally, we elucidate the benefits of cross-modal ensembling from the perspective of model \emph{diversity}, a key characteristic of effective ensembles~\citep{kuncheva2003measures}. 
We explore two simple facets of diversity between pairs of models in an ensemble.
The first is \textbf{rank correlation}, or the Pearson correlation coefficient of the gold answer ranks between two models. 
% We report the rank correlation on \repodb{}, \codexS{}, and \codexM{} between the best and second-best \kge{}s, the best \kge{} and the StAR dual-encoder, and the best \kge{} and the KG-BERT cross-encoder in 
As shown in Table~\ref{table:rank-correlation}, rank correlation is highest between pairs of \kge{}s and lowest between the \kge{} and cross-encoder.
This observation corresponds well with ensemble performance, as we find that the most ``diverse'' ensembles in terms of rank correlation (i.e., those that combine a \kge{} + cross-encoder) are the most competitive. 

\begin{table}[!t]
    \caption{
    The ranks of gold answers are least correlated (Pearson's correlation coefficient, all $p$-values 0) between the \kge{} and the cross-encoder on the dev set, suggesting that these two model types provide the most diverse or complementary link prediction performance.
    % Ranks are computed on the validation set. 
    % Correlation is reported with Pearson's correlation coefficient. All $p$-values are 0. 
    % We omit \fb{} and \wn{} because full cross-encoding was too costly on these datasets. 
    }
    \label{table:rank-correlation}
    \centering
    \resizebox{0.55\textwidth}{!}{
        \begin{tabular}{l cccc} 
        \toprule
        & (\kge{}, \kge{}) & (\kge{}, dual-enc.) & (\kge{}, cross-enc.) \\
        \toprule 
        \repodb{} & 0.7371 & 0.3494 & 0.2025 \\ 
        \codexS{} & 0.8081 & 0.7265 & 0.6163 \\
        \codexM{} & 0.6471 & 0.5406	& 0.4865 \\ 
        % \wn{} \\ 
        \bottomrule
        \end{tabular}
    }
\end{table}

\begin{figure}[!t]
    \centering
    \begin{subfigure}{0.3\columnwidth}
        \includegraphics[width=\textwidth]{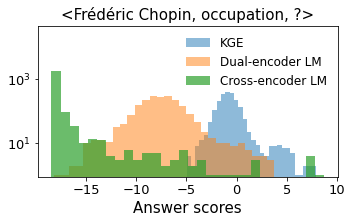}
    \end{subfigure}
    ~
    \begin{subfigure}{0.3\columnwidth}
        \includegraphics[width=\textwidth]{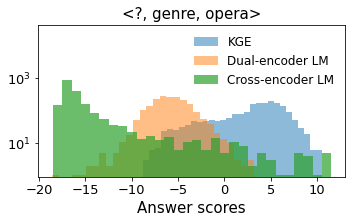}
    \end{subfigure}
    \caption{
    The cross-encoder's score distributions are skewed left compared to the \kge{}'s and dual-encoder's score distributions. 
    Shown are two randomly-selected link prediction queries from \codexM.
    The trends in the plots hold across queries and datasets. 
    }
    \label{fig:score-distributions}
\end{figure}

Next, we consider each model's empirical \textbf{score distributions} to two queries in Figure~\ref{fig:score-distributions}. 
The figure demonstrates that the cross-encoder's score distributions are skewed left compared to those of the \kge{} and the dual-encoder.
Moreover, we find that this trend holds across queries and datasets, suggesting a fundamental difference in scoring behavior. 
We hypothesize that cross-encoders may  filter out irrelevant candidate answers to queries more aggressively, perhaps because they model term overlap between text pairs with relatively high precision compared to dual-encoder models that do not use cross-attention~\citep{luan-etal-2021-sparse}.
Again, this finding corresponds well with ranking performance, as the model pairs with more diverse score distributions (i.e., \kge{}s and cross-encoders) make stronger cascades. 

\begin{figure}[!t]
    \centering
    \begin{subfigure}{0.47\columnwidth}
        \includegraphics[width=\textwidth]{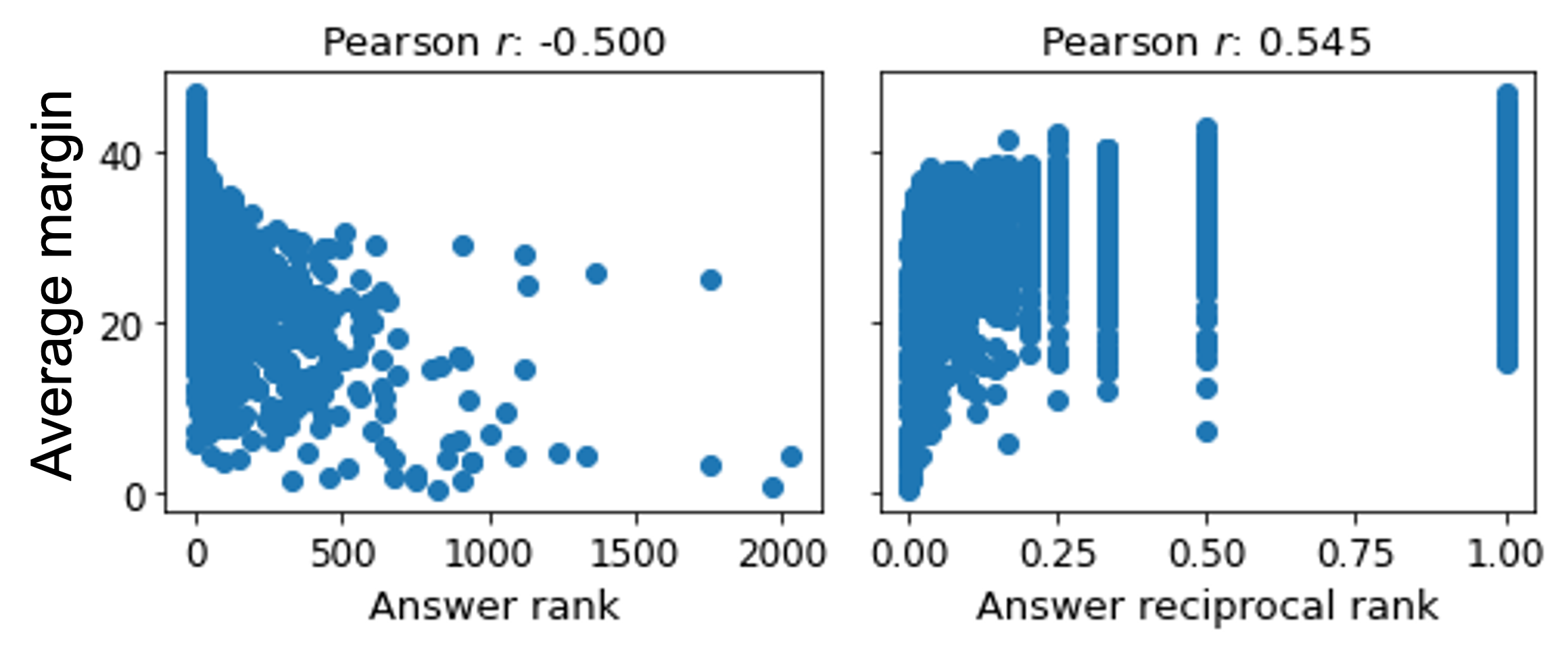}
        \label{fig:gaps-no-weighting}
        \caption{KGE + cross-enc. without reweighting}
    \end{subfigure}
    ~
    \begin{subfigure}{0.47\columnwidth}
        \includegraphics[width=\textwidth]{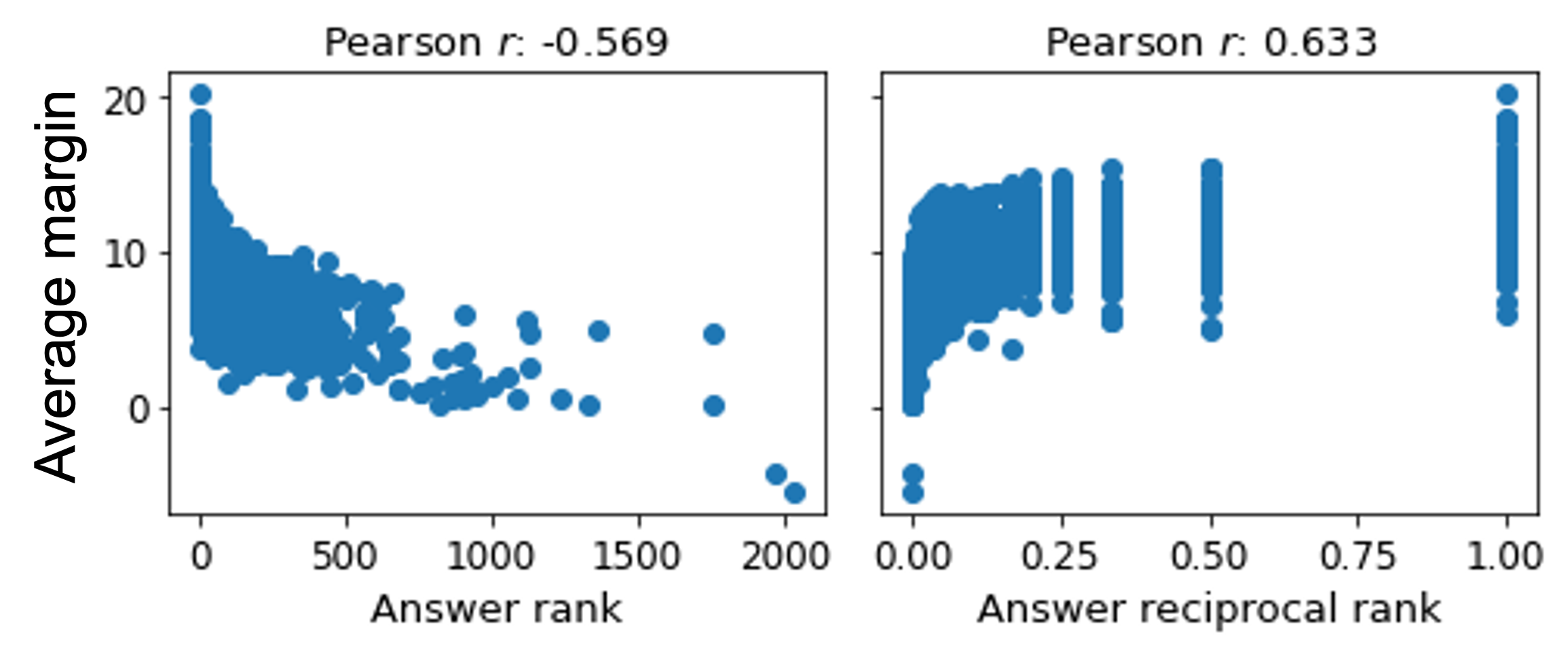}
        \label{fig:gaps-weighting}
        \caption{KGE + cross-enc. with reweighting}
    \end{subfigure}
    \caption{
    Ensembling with additive reweighting preserves the correlation between gold answer ranks and average margins.
    }
    \label{fig:score-gaps}
\end{figure}

Finally, having established the importance of diversity, we answer the question: 
When ensembling highly diverse models, how crucial is additive reweighting as defined in \eqref{eq:additive-reweighting}, versus simply summing the models' individual scores without reweighting? 
Across all datasets, we observe \cascader{}'s dev MRR drops 6-8 points without reweighting (see Appendix~\ref{appendix:additional-results} for exact comparisons). 
% which compares the MRR of the best \cascader{} architectures with and without additive reweighting, performance drops significantly without reweighting. 
To explain this phenomenon, we investigate the \textbf{average margin} between the gold answer and all negative candidates for a query, i.e., for query $i$ with gold tail entity $j^{+}$ and $N^{-}$ negative candidates $j^{-}$,
the average margin is defined as
$\frac{1}{N^{-}} \sum_{j^{-}} S_{ij}^{+} - S_{ij}^-$.
In Figure~\ref{fig:score-gaps}, we plot the gold ranks and reciprocal ranks of answer entities on \codexM{} against their average margins with and without reweighting a \kge{} + cross-encoder ensemble. 
We observe that the correlation between margins and gold ranks is higher under additive reweighting, which suggests that reweighting helps preserve the confidence signals in the base models' margins, whereas ensembling without reweighting dilutes these signals.
% Further, we find that normalizing score outputs tended to also dilute these confidence signals, reducing ensemble accuracy. 

% \section{Related Work}
% \label{sec:related}
% \input{05related}

\section{Conclusion}
\label{sec:conclusion}
In this paper, we considered the task of \kg{} link prediction with cross-modal ensembles. 
Motivated by the inherent accuracy-efficiency tradeoff, we proposed \cascader{}, a novel cross-modal reranking architecture that uses deep language models to rerank the outputs of knowledge graph embeddings. 
We showed that \cascader{} achieves state-of-the-art performance on multiple link prediction benchmarks by effectively combining structure and text, while improving efficiency over our strongest ensemble baseline by orders of magnitude. 

Our work opens up several avenues for future research. 
For one, more advanced candidate pruning strategies may further increase the efficiency of \cascader{} while maintaining accuracy.
For another, ``hybrid'' \lm{}s that attempt to interpolate between the efficiency of dual-encoders and the effectiveness of cross-encoders~\citep{khattab2020colbert,luan-etal-2021-sparse} may improve \cascader{}'s ability to balance these two desiderata.
Finally, extensions of \cascader{} may help solve the inductive link prediction, in which novel entities/relations are presented at test time~\citep{galkin-etal-2022-inductive}, as contextual \lm{}s are naturally inductive and can correct the shortcomings of transductive \kge{}s in this setting.

% \acks{
% We thank Hannaneh Hajishirzi and Danai Koutra for stimulating discussions and feedback.  This work was supported in part by NSF Convergence Accelerator Award \#2132318.
% }

\bibliographystyle{plainnat}
\bibliography{references}

\clearpage

\appendix
\section{}
\subsection{Additional related work}
\label{appendix:related-work}
\paragraph{Joint modeling on \kb{}s}
The task of inferring novel links in \kb{}s has been widely studied. 
The most prevalent approaches are structure-only \kb{} embeddings~\citep{nickel-etal-2011-three,bordes-etal-2013-translating,trouillon-etal-2016-complex,sun-etal-2019-rotate,balazevic-etal-2019-tucker,ji-etal-2020-survey}. 
That said, prior to the advent of pretrained contextual \lm{}s, a few cross-modal structure and text approaches for link prediction were  proposed~\citep{toutanova-etal-2015-joint,toutanova-etal-2016-compositional,xie-etal-2016-entity-descriptions}.
Such approaches rely on convolutional text representation architectures to obtain embeddings of entities or relations using, e.g., entity descriptions~\citep{xie-etal-2016-entity-descriptions} or textual relation mentions~\citep{toutanova-etal-2015-joint}.
These text-based embeddings are then composed using  structural \kb{} embedding scoring functions to score novel links in the \kb. 

More recently, motivated by the successes of Transformer language models, 
pretrained Transformer \lm{}s like BERT~\citep{devlin-etal-2019-bert} have begun to gain traction for variants of the link prediction task~\citep{yao-etal-2019-kg-bert,kim-etal-2020-multi,daza-etal-2021-inductive,wang-etal-2021-structure-augmented,nadkarni-etal-2021-scientific}. 
The approaches most related to \cascader{} are the ensembles considered by~\citet{wang-etal-2021-structure-augmented} and~\citet{nadkarni-etal-2021-scientific},
both of which construct additive ensembles of structural \kb{} embeddings and contextual \lm{}s.
Compared to these baseline ensembles, \cascader{} improves accuracy and/or efficiency over both. 

\paragraph{Cascade models}
Multi-stage cascade ensembles have been successful in computer vision~\citep{viola-jones-2001-cascade,wang-etal-2022-wisdom-committees} and text retrieval~\citep{wang-etal-2011-cascade,chen-etal-2017-cost-aware,gallagher-etal-2019-joint-optimization,lin-etal-2021-text-ranking}. 
Recently, several studies have proposed to use BERT as a late-stage ranker in multi-stage document retrieval~\citep{nogueira-etal-2019-multi-stage} and passage retrieval~\citep{matsubara-etal-2020-reranking} pipelines.
Similar to our work, these studies are motivated by the observation that using BERT in a multi-stage cascaded setting can significantly boost retrieval accuracy while maintaining efficiency~\citep{lin-etal-2021-text-ranking}. 
Yet other studies have attempted to balance the effectiveness-efficiency tradeoff by proposing dual-encoding architectures that are more efficient but usually less effective than cross-encoder BERT models for information retrieval~\citep{reimers-gurevych-2019-sentence,humeau-etal-2020-poly-encoders,karpukhin-etal-2020-dense,xiong-etal-2020-ance,khattab2020colbert}.
Our work builds upon all of these important insights, which have been instrumental in scaling contextual \lm{}s to large-scale text ranking. 
As far as we are aware, we are the first to bridge these ideas with the traditional graph learning task of link prediction.

\subsection{Model selection}
\label{appendix:model-selection}

We implement all \kg{} embeddings using the open-source Lib\kge{} PyTorch library~\citep{broscheit-etal-2020-libkge}.
We use the pretrained  \kge{} checkpoints provided by LibKGE for \fb{} and \wn{}.
For the other datasets, we follow a similar hyperparameter tuning strategy to that proposed by~\citet{ruffinelli-etal-2020-you} for tuning our \kge{} baselines.

We implement all \lm{}s with the Huggingface transformers PyTorch  library~\citep{wolf-etal-2020-transformers} using the same base language model, which is BERT-\textsc{Base}~\citep{devlin-etal-2019-bert} for all benchmarks except \repodb{}, and \textsc{PubMedBERT}~\citep{gu-etal-2021-domain} for \repodb.
We use the following hyperparameters:
Batch size of 16, 
learning rate of $10^{-5}$,
and 10 epochs.
We use a maximum sequence length of 32, 64, and 256 respectively for \wn{}, \repodb{}, and all other datasets. 
For the dual-encoder \lm{} we use 16 negative samples per positive.
For the cross-encoder \lm{} we use 2 negative samples per positive. 

For all of the ensemble baselines and \cascader{}, we tune the weighting hyperparameter $\alpha \in [0.05, 0.95]$. 
All experiments are conducted on a single NVIDIA Quadro RTX 8000 GPU with 48 GB of RAM. 
All main results reported in the paper use a three-tiered structure with no pruning between tiers one and two and dynamic pruning at $q=0.9$ between tiers two and three.
We provide exact wall-clock inference time for \cascader{} in Table~\ref{table:timing-details}. 

\begin{table*}[t!]
\centering
\caption{Wall-clock inference time of \cascader{} using a single RTX 8000 GPU. 
}
\label{table:timing-details}
\resizebox{0.4\textwidth}{!}{
    \begin{tabular}{ l l }
    \toprule
    & Inference time total \\ 
    \toprule 
    \fb{} & 20 hours \\
    \wn{} & 3.5 hours \\ 
    \codexS{} & 26 min \\
    \codexM{} & 17 hours \\
    \repodb{} & 5 min\\
    \bottomrule
\end{tabular}
}
\end{table*}

\subsection{Additional results}
\label{appendix:additional-results}

\paragraph{Results on \codexS{}}
Table~\ref{table:codexs-results} provides link prediction performance on the \codexS{} dataset~\citep{safavi-koutra-2020-codex}.
\cascader{} again achieves state-of-the-art performance on this dataset, suggesting its wide applicability across \kg{}s. 

\paragraph{Importance of additive reweighting}
Table~\ref{table:reweighting} compares performance of a \kge{} + cross-encoder ensemble with and without additive reweighting, as defined in Eq.~\eqref{eq:additive-reweighting}.
We observe that MRR drops 6-8 points without the reweighting term, confirming that additive reweighting is key to preserving the base models' confidence signals in the ensemble.

 \begin{table}[!t]
    \caption{
    \cascader{} achieves state-of-the-art test performance on \codexS{}.
    \textbf{\underline{Bold + underline}}: Best performance.
    \underline{Underline}: Second-best performance. 
    }
    \label{table:codexs-results}
    \centering
    \resizebox{0.6\textwidth}{!}{
        \begin{tabular}{l cccc }
        \toprule
        & 
        MRR & H@1 & H@3 & H@10 \\ 
        \midrule 
        RESCAL & 0.4040 & 0.2935 & 0.4494 & 0.6225 \\ 
        TransE & 0.3540 & 0.2185 & 0.4218 & 0.6335 \\
        ComplEx & 0.4646 & 0.3714 & 0.5038 & 0.6455 \\ 
        RotatE & 0.2587 & 0.1586 & 0.2916 & 0.4609 \\ 
        \midrule 
        StAR & 0.3540 & 0.2306 & 0.4051 & 0.6007 \\ 
        KG-BERT & 0.2849 & 0.1472 & 0.3310 & 0.5848 \\ 
        \midrule 
        \kge{} + \kge{} & 0.4665 & 0.3712 & 0.5082 & 0.6518 \\ 
        \kge{} + StAR & 0.4751 & \underline{0.3717} & 0.5249 & 0.6712 \\ 
        \kge{} + KG-BERT & \underline{0.4812} & \textbf{\underline{0.3764}} & \underline{0.5290} & \textbf{\underline{0.6898}} \\ 
        \midrule 
        SOTA & 0.4646 & 0.3714 & 0.5038 & 0.6455 \\ 
        \midrule 
        \textbf{\cascader{}} & \textbf{\underline{0.4839}} & \textbf{\underline{0.3764}} & \textbf{\underline{0.5383}} & \underline{0.6871} \\ 
        \bottomrule
        \end{tabular}
    }
\end{table}

\begin{table}[!t]
    \caption{
    Additive reweighting is crucial to ensembling \kge{}s and \lm{}s. 
    We compare the MRR of the best \cascader{} architecture \emph{with} additive reweighting as defined in Eq~\eqref{eq:additive-reweighting} to the best \cascader{} architecture without the reweighting term.
    }
    \label{table:reweighting}
    \centering
    \resizebox{0.5\textwidth}{!}{
        \begin{tabular}{l cc} 
        \toprule
        & No reweighting & Reweighting \\ 
        \toprule 
        \repodb{} & 0.4498 & 0.5156 \\ 
        \codexS{} & 0.4204 & 0.4839 \\
        \codexM{} & 0.3114 & 0.3830 \\ 
        % \wn{} \\ 
        \bottomrule
        \end{tabular}
    }
\end{table} 

\end{document}